\RequirePackage{snapshot}       
\documentclass[english]{spie}
\usepackage[%
 hidelinks,%
  pdftitle={Population-based Respiratory 4D Motion Atlas Construction and its Application for VR Simulations of Liver Punctures},%
  pdfauthor={Andre Mastmeyer, Matthias Wilms, Heinz Handels},%
  pdfsubject={My Subject},%
  pdfkeywords={Virtual Reality, Liver Puncture Training, 4D Motion Models,Application of 4D Motion Models}]{hyperref}
\usepackage[pscoord]{eso-pic}
\usepackage[color=black,opacity=1,angle=0,scale=1]{background}
\newcommand{\placetextbox}[3]{
  \setbox0=\hbox{#3}
  \AddToShipoutPictureFG*{
    \put(\LenToUnit{#1\paperwidth},\LenToUnit{#2\paperheight}){\vtop{{\null}\makebox[0pt][c]{#3}}}%
  }%
}%

\backgroundsetup{
  contents={%
    \placetextbox{0.5}{0.1}{To appear in SPIE Medical Imaging: Image Processing 2018}
    \placetextbox{0.5}{0.085}{p. \thepage}%
  }
}
\usepackage[T1]{fontenc}
\usepackage[latin9]{inputenc}
\usepackage{graphicx}
\usepackage{amsmath}
\usepackage{amssymb}
\usepackage{bm}
\usepackage{color}

\makeatletter

\title{Population-based Respiratory 4D Motion Atlas Construction and its Application for VR Simulations of Liver Punctures}

\author{Andre Mastmeyer, Matthias Wilms, and Heinz Handels
\skiplinehalf
Institute of Medical Informatics, University of L\"ubeck, L\"ubeck, Germany\\
}

\authorinfo{Further author information: (Send correspondence to Andre Mastmeyer; E-mail: mastmeyer@imi.uni-luebeck.de). This work is supported by the German Research Foundation: DFG HA 2355 / 11-2.}

\@ifundefined{showcaptionsetup}{}{%
 \PassOptionsToPackage{caption=false}{subfig}}
\usepackage{subfig}
\makeatother

\usepackage{babel}
\begin{document}
\maketitle
\begin{abstract}
Virtual reality (VR) training simulators of liver needle insertion in the hepatic area of breathing virtual patients often need 4D image data acquisitions as a prerequisite.
Here, first a population-based breathing virtual patient 4D atlas is built and second
the requirement of a dose-relevant or expensive acquisition of a 4D CT or MRI data set for a new patient can be mitigated by warping the mean atlas motion. The breakthrough contribution of this work is the construction and reuse of population-based, learned 4D motion models.
\end{abstract}

\section{Purpose}
\begin{figure}[t]
  \centering
    \subfloat[]{\includegraphics[bb=0bp 0bp 1728bp 2592bp,clip,height=5.8cm]{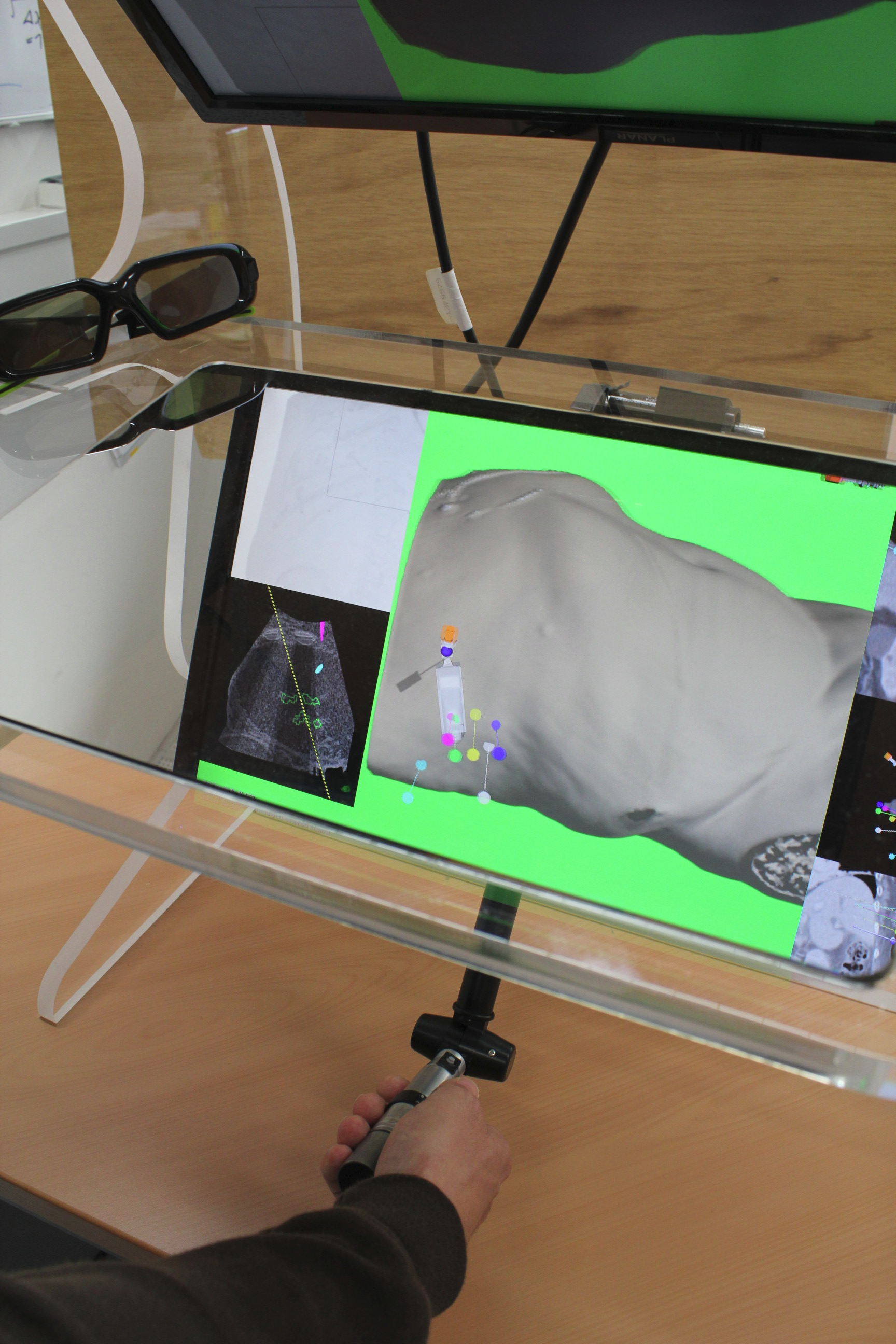}}
	 \quad \subfloat[]{\includegraphics[bb=0bp 0bp 1524bp 970bp,clip,height=5.8cm]{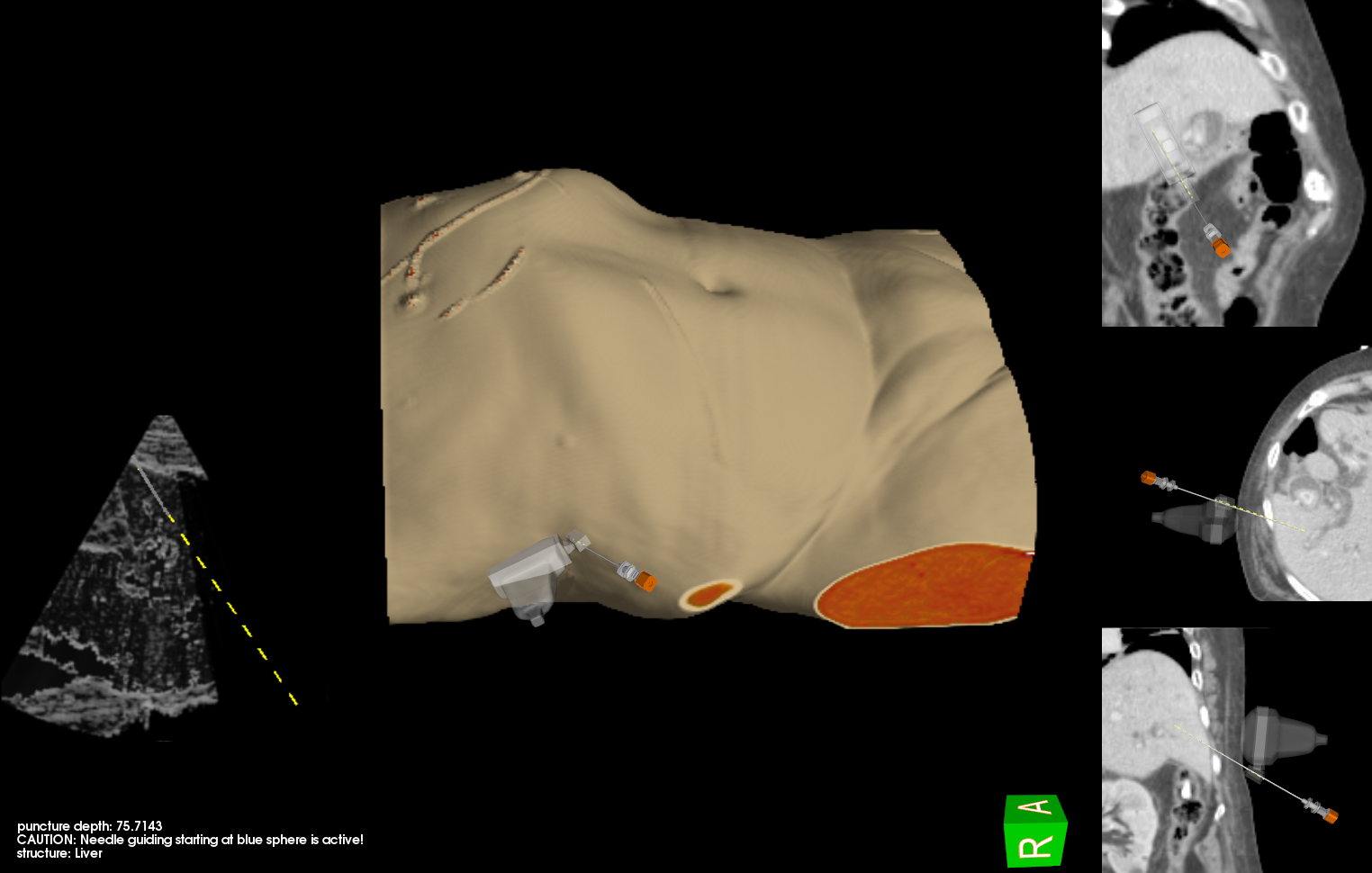}}
\caption{\label{SPIE18-fig1}VR simulator setup: (a) Semi-transparent rendering mirror with successful needle insertion indicated by green background, workbench in which a stereo-monitor is mounted above the mirror, haptic steering device below the mirror with device handle held by the needle steering user. (b) Graphical user interface with breathing virtual patient and assisting viewports, e.g. ultrasound simulation in the bottom left viewport.}
\end{figure}

In recent works, for the risk-free training and planning of surgical needle interventions with visuo-haptic virtual reality simulators the inclusion of breathing motion models \cite{McClelland2013} is a core component.
The simulation of tissue and needle deformation for puncture interventions and the modelling of virtual patient bodies \cite{mastmeyer2017accurate,Mastmeyer2016RFC_LVS,meike2015real,SPIE18-09,SPIE18-10,mastmeyer2012anisotropic,engelke2006effect} has been an active research branch for years 
\cite{SPIE18-01,SPIE18-04,mastmeyer2013ray,SPIE18-05,SPIE18-12,Fortmeier2013a}
and breathing motion can be rendered in a recent 4D visuo-haptic simulator (Fig. \ref{SPIE18-fig1}) with GPU support using 4D direct volume rendering \cite{mastmeyer2017evaluation,Mastmeyer2016MMVRUSSimu,SPIE18-01a,FortmeierMH13,fortmeier2012gpu,Mastmeyer2012}.

In our setup \cite{SPIE18-01a}, the visuo-haptic simulation of the patient's breathing motions is a new key feature. \cite{SPIE18-04,SPIE18-05,fortmeier2012gpu}. 
In the abdominal liver region, respiratory movements dominate the visuo-haptic experience of the physician with possible breathing displacements of up to 5 cm. \cite{SPIE18-07}. 
Conceptually, with a 4D breathing model, a new 3D CT data set could be animated. The model can be patient-specific\cite{mastmeyer2017interpatient} or an averaged population-based model, 
which is the main topic of this work. Such a model then can be warped by non-linear registration to the 3D CT data set of a new patient. \cite{mastmeyer2017interpatient}.

Surrogate free mean motion models were proposed by our group \cite{SPIE18-16} for the lungs. The motivation for such models is to supply a more general model by averaging out patient specific breathing patterns. Thus,  transferring such models instead of a patient-specific model to a new static 3D patient can be favorable. 

Typically for VR training purposes, the acquisition of new 4D CT or MR data is impossible for ethics and cost reasons. Thus the transfer of existing motion models is appealing.

Here, we present a first study of an effective and efficient building process of a mean 4D breathing atlas parametrized by a surrogate signal and apply it to 3D patient data for rendering it in a 4D VR environment \cite{SPIE18-01a}.

\section{Materials and Methods}

\subsection{Study Data\label{sub:Data-and-Preprocessing}}

The 4D mean model is built from $N_{pats_{4D}}=5$ 4D-CT data sets (Pat. 1-5, Fig. \ref{SPIE18-img:FoV}a-c) of the thorax and upper abdomen. Corresponding $N_{phases}=14$ respiratory phase images (up to $N_{vox}=$512x512x462, 1 mm$^3$) with low dimensional spirometry parameter signals $ v^p (t) $ were used (Fig.~\ref{SPIE18-img:FoV}d).
The new patient is represented only by a static 3D CT data set in a frame of reference (e.g. reference phase from Pat. 3).
Selected reference phases $ j_{\mathrm{ref}} $ correspond to phase images in a corresponding inhalation state (new 3D patients hold breath, Fig.~\ref{SPIE18-img:FoV}a-c).

\subsection{Intra-patient Motion Modeling in a Frame of Reference}
\label{SPIE18-sec:model}
Breathing motion modeling relies on $N_{pats_{4D}}$ 4D CT data sets with $N_{phases}$ phases, indexed by $j\in\{1..N_{phases}\}$ and $p\in\{1..N_{pats_{4D}}\}$. Spirometry signals $v^p (t)$ [ml] serve as surrogate signal components to represent the patient's breathing state: $z^p(t)=(v^p (t), v'^p(t))$.
The derivative $v'^p (t)$ allows to model respiratory hysteresis (inhalation vs. exhalation). Signal and motion dependence are assumed (locally) linear. $N_{phases}-1$ intra-patient inter-phase image registrations to a selected patient specific reference phases $j_{ref}$ are conducted first:
\begin{eqnarray}
\label{eq:intra}
\varphi^{p}_{j \rightarrow j_{ref}}=
\underset{\varphi}{\mathrm{argmin}}\left(D_{intra}\left[I^{p}_{j_{ref}}, I^{p}_{j} \circ \varphi\right]
+\alpha \cdot R_{intra}(\varphi)\right),
\\ \nonumber j\in \{1,..,\mathrm{N}_{phases}\},~j\neq j_{ref}
\\ \nonumber p\in \{1,...,N_{pats_{4D}}\}.
\end{eqnarray}
Using the distance metric $D_{intra}$ and regularizer $R_{intra}$, this step yields motion fields for each 4D patient data set. These can be used to calculate vector field coefficients $ \hat{A}=(a^{p}_{1},a^{p}_{2},a^{p}_{3})$ for the application of a linear multivariate regression model. 

Let us introduce a bijective serialization operator $\mathrm{ser}(.)$ that produces a long column vector in a standardized way. The serialization of $\hat A$ is $A \in \mathbb{R}^{3\cdot N_{vox}\times 3}$. For one patient's reference space, the serialized vector fields as regressand $V=\mathrm{ser}(\varphi_{1\rightarrow j_{ref}},...,\varphi_{N_phases\rightarrow j_{ref}})\in \mathbb{R}^{3\cdot N_{vox}\times N_{phases}}$ are approximated by linear regression subject to an expanded serialized surrogate signal with $z_1(\mathrm t)=(v^p (\mathrm t), v'^p(\mathrm t),1)  \in \mathbb{R}^{3}$, i.e. $Z=\mathrm{ser}(z_1(\mathrm t))\in \mathbb{R}^{3 \times N_{phases}}$ as regressor and discrete time points $\mathrm t=j\in{1..N_{phases}}$ \cite{wilms2017}:
\begin{equation}
\hat{A}=\mathrm{ser}^{-1}\left(\underset{A}{\mathrm{argmin}} ||V-A \cdot Z||^{2}_{F}\right)
\end{equation}
where $F$ denotes the Froebenius norm \cite{hastie2009elements}.

We yield a motion estimate using the solution $\hat{A}$ for any point in time:
\begin{eqnarray}\label{eq:modelIntra}
\hat{\varphi}^{p}(\mathbf{x},t)=
a^{p}_1(\mathbf{x})\cdot v^p(t)~+
a^{p}_2(\mathbf{x})\cdot v'^p(t)~+
a^{p}_3(\mathbf{x}),~~\mathbf{x} \in \Omega_{p}.
\end{eqnarray}
Now, we can simulate each 4D patient's breathing state over time $t$: 
\begin{equation}
I^{p}(t)=I^{p}_{j_{\mathrm{ref}}} \circ \hat{\varphi}^{p}(\mathbf{x},t)
\end{equation}
Our latest VR training simulator can render this compact, personalized representation of a breathing virtual patient $I^{p}(t)$ in realtime by efficient raycasting using a bent rays approach \cite{SPIE18-01a}.

\subsection{Population-based Breathing Motion Models}
The next step is the averaging of the personalized breathing models to   yield a mean 4D motion model.

For a common frame of reference image phase, a reference patient $p_{\mathrm{ref}}$ is selected from the 4D patient population. This patient is targeted by  inter-patient registrations in the reference phase $j_{ref}$. 

Nonlinear registrations $ \varphi (\mathbf {x}): \Omega_{p} \rightarrow \Omega_{p_{\mathrm{ref}}} $, minimizing the distance metric $ D_{inter} $, warp the patient's image data to the selected reference patient $p_{\mathrm{ref}}$:
\begin{eqnarray}\label{eq:interAvg}
\varphi^{p \rightarrow p_{\mathrm{ref}}}_{j_{\mathrm{ref}}}=
\underset{\varphi}{\mathrm{argmin}}\left(D_{inter}\left[I^{p_{\mathrm{ref}}}_{j_{\mathrm{ref}}}, I^{p}_{j_{\mathrm{ref}}} \circ \varphi\right]
+\beta \cdot R_{inter}(\varphi)\right), p\neq p_{\mathrm{ref}},
\end{eqnarray}
using a distance measure $ D_{inter} $, regularizer $R_{inter}$, and $\beta$ a regularizer weight.

Next by $\varphi^{p \rightarrow p_{\mathrm{ref}}}_{j_{\mathrm {ref}}}$ , we warp the intra-patient-inter-phase deformations $\varphi^{p}_{j} $ of the 4D patients to the common reference frame $j_{\mathrm{ref}}$ of $p_{\mathrm{ref}}$:
\begin{equation}\label{eq:backNForthWarpAvg} 
{\varphi}^{p_{\mathrm{ref}}}_{j,p} = 
\varphi^{p \rightarrow p_{\mathrm{ref}}}_{j_{\mathrm{ref}}} 
\circ 
\varphi^{p}_{j} \circ
\left( 
\varphi^{p \rightarrow p_{\mathrm{ref}}}_{j_{\mathrm{ref}}}\right)^{-1}.
\end{equation}
With all motion fields in the same common frame of reference, averaging yields the prerequisite for the mean motion model:
\begin{equation}\label{eq:avgFields} 
\hat{\varphi}^{avg}_{j}(\mathbf{x})=
\frac{1}{N_{pats_{4D}}} \sum_{p=1}^{N_{pats_{4D}}} \varphi^{p_{\mathrm{ref}}}_{j,p}
\end{equation}
Averaging is also done for the surrogate signals: $v^{avg}(t)=1/N_{pats_{4D}} \sum_{p=1}^{N_{pats_{4D}}}v^p(t) $.
Finally, the mean patient's motion behavior can now be estimated efficiently similar to one of the 4D patients (see section \ref {SPIE18-sec:model}):
\begin{equation}\label{eq:modelAvg} 
\hat{\varphi}^{avg}(\mathbf{x},t)=
{a}^{avg}_1(\mathbf{x})\cdot v^{avg}(t)~+
{a}^{avg}_2(\mathbf{x})\cdot v'^{avg}(t)~+
{a}^{avg}_3(\mathbf{x}),~~\mathbf{x} \in \Omega_{p_{\mathrm{ref}}}.
\end{equation}
Image data for the artificial mean patient $I^{avg}_{j_{\mathrm{ref}}}$ underwent an analogous transformation and averaging pipeline as the fields.

\subsection{Application of Mean Motion Models} \label{SPIE18-sec:applications}

Three applications are at hand now at this juncture:

\begin{enumerate}
\item An individual patient from the training population can be animated by the mean motion model and be presented to the user for training purposes (Fig. \ref{SPIE18-img:motionct}a-d). This is alternative to using the patient-individual motion model \cite{mastmeyer2017interpatient,SPIE18-01a} providing a less individual or situational breathing pattern. In radiation therapy, this could make sense in follow-up fractional treatment sessions.

\item The population based patient intensity and motion atlas can be used to present an interactively manipulable breathing virtual training atlas to the apprentice user (Fig. \ref{SPIE18-img:motionct}e-h).

\item The most relevant application of the 4D mean patient breathing model is warping it to new unseen 3D static patient data and render it in our current 4D VR simulator (Fig. \ref{SPIE18-img:motionct}i-l).
\end{enumerate}
Details of the inter-patient model transfer can be found in \cite{mastmeyer2017interpatient}. 
To this aim, we restate Eq. \ref{eq:interAvg}:
\begin{equation}\label{eq:interNew}
\varphi^{avg \rightarrow new}_{j_{\mathrm{ref}}}=\\ 
\underset{\varphi}{\mathrm{argmin}}\left(D_{inter}\left[I^{new}_{j_{\mathrm{ref}}}, I^{avg}_{j_{\mathrm{ref}}} \circ \varphi\right]
+\beta \cdot R_{inter}(\varphi)\right),
\end{equation} 
and Eq. \ref{eq:backNForthWarpAvg} accordingly to yield the new patient's motion behaviour:
\begin{equation}
{\varphi}^{new}_{j} = 
\varphi^{avg \rightarrow new}_{j_{\mathrm{ref}}} 
\circ 
\varphi^{avg}_{j} \circ 
\left(
\varphi^{avg \rightarrow new}_{j_{\mathrm{ref}}}
\right)^{-1}.
\end{equation}
$\hat{\varphi}^{new}(\mathbf{x},t)$ then can analogously be built as in Eqs. \ref{eq:modelIntra} and \ref{eq:modelAvg}.

The respiratory motion applied to the new 3D patient image data can now be calculated efficiently (see section \ref {SPIE18-sec:model}). Thus based only on a relatively low dose 3D-CT data acquisition in reference phase $I^{new}_{j_{\mathrm{ref}}}$ and the transferred mean motion model, the breathing movements can be plausibly approximated:
\begin{equation}\label{eq:model3D}
\hat{\varphi}^{new}(\mathbf{x},t)=
{a}^{avg}_1(\mathbf{x})\cdot v(t)~+
{a}^{avg}_2(\mathbf{x})\cdot v'(t)~+
{a}^{avg}_3(\mathbf{x}),~~\mathbf{x} \in \Omega_{new}.
\end{equation}
and animated:
\begin{equation}
I^{new}(t)=I^{new}_{j_{\mathrm{ref}}} \circ \hat{\varphi}^{new}(\mathbf{x},t).
\end{equation}

Optionally, simulated surrogate signals $ v (t) $ with stochastic irregularity \cite{wilms2014} in the range of the observed breathing states 
can be used for the 4D animation of the mean atlas and new 3D CT data by such a mean motion model. In the data sets used here, we found relative breathing volumes of ca. 0 - 1200 ml.

\begin{figure}
\center
\subfloat[]{\includegraphics[bb=0bp 0bp 417bp 518bp,clip,height=5cm]{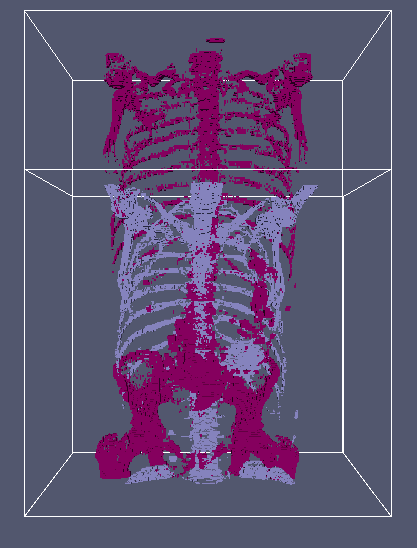}}\hspace{.1mm}
\subfloat[]{\includegraphics[bb=0bp 0bp 417bp 518bp,clip,height=5cm]{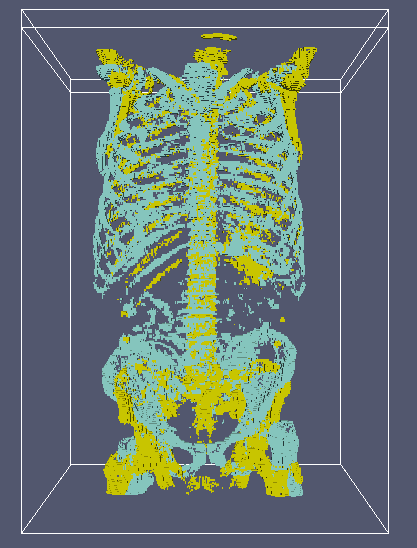}}\hspace{.1mm}
\subfloat[]{\includegraphics[bb=0bp 0bp 417bp 518bp,clip,height=5cm]{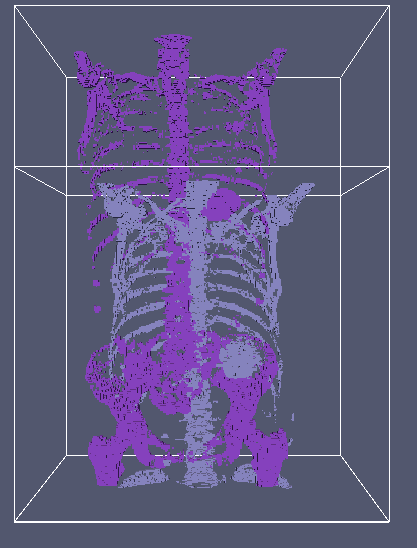}}\hspace{.1mm}\subfloat[]{\includegraphics[bb=2bp 2bp 300bp 276bp,clip,height=4.4cm]{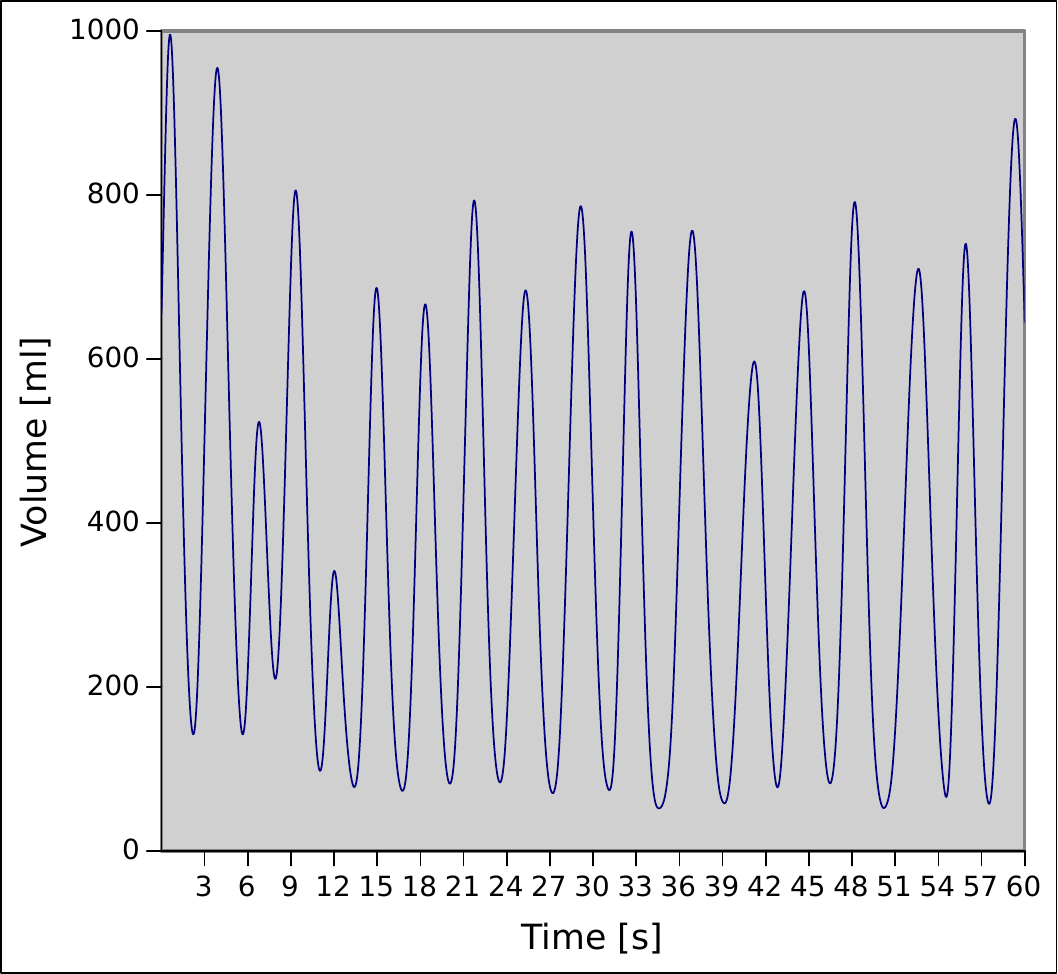}}
\caption{Field of view differences of the patient image data: (a) Pat. 1 (lilac), Pat. 2 (wine red). (b) Pat. 3 (yellow), Pat. 4 (turquois). (c) Pat. 5 (purple), Pat. 1 (lilac). (d) Example spirometry signal from Pat. 3.
}
\label{SPIE18-img:FoV}
\end{figure}

\subsection{Study Set-up}
Volume image data was resampled to a size of 256$^3$ voxels to fit into moderate graphics cards with 2 GB GPU-RAM (Nvidia GTX 680) also comprising the three vector field coefficients $a_1..a_3$ \cite{SPIE18-01a}.
From our 4D patient population with fields of view shown in Fig. \ref{SPIE18-img:FoV}, for one experiment a representative 4D-CT data set $p_{\mathrm{ref}}$ in maximum inhalation phase $j_{\mathrm{ref}}$ of the thorax and upper abdomen serves as common reference frame and is left out from the motion model building process.
The new patient for application option three is represented only by a static 3D CT data set. The reference patient with the image data $I^{p_{\mathrm{ref}}}_{j_{\mathrm{ref}}}$ and the new 3D patient image data $I^{new}_{j_{\mathrm{ref}}}$ are assumed to be in the same breathing state. 
For each of the $N_{pats_{4D}}$ 4D patients and according to Eq.~\ref{eq:intra}, we first perform the intra-patient inter-phase registrations to the chosen reference phase $j_{ref}$ of the currently considered 4D image data.
In Eq. \ref{eq:intra}, we use the distance measure $ D_{intra}=D_{NSSD} $ (normalized sum of voxel-wise squared differences 
) and $R_{intra}=R_{SMP} $ (sliding motion'-preserving) to cope with discontinuities in the pleural cavity \cite{SPIE18-14}. 
In Eqs. \ref{eq:interAvg} and \ref{eq:interNew}, $ D_{inter} = D_{SSD}$ (sum of squared differences) and $ R_{inter}=R_{DNL} $ (diffusive non-linear regularization). Regularization weights are set as $\alpha=0.1$ or $\beta=1.0$, as interphase registration requires lesser regularization influence (same anatomy morphology).
DICE coefficients of warped expert segmentation masks (liver, left/right lungs) from a leave-one-out crossvalidation are given to classify the quality of the difficult inter-patient registration. We present sample images for the three use cases, four time instants and online movie footage for reference patient 3 and the spirometry signal from Fig. \ref{SPIE18-img:FoV}d in application 2 (mean intensity and motion atlas). 

\begin{figure*}
  \centering
  	\subfloat[0 ml\label{sfig:}]{\includegraphics[bb=0bp 0bp 315bp 511bp,clip,width=3.7cm]{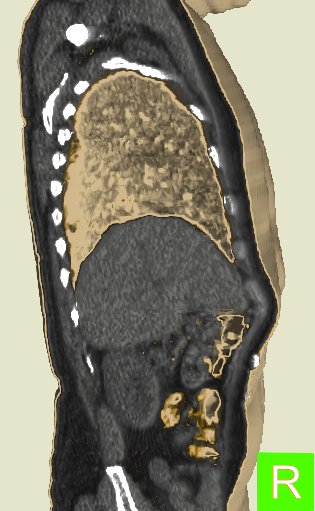}}\qquad		    
    \subfloat[349 ml\label{sfig:}]{\includegraphics[bb=0bp 0bp 315bp 511bp,clip,width=3.7cm]{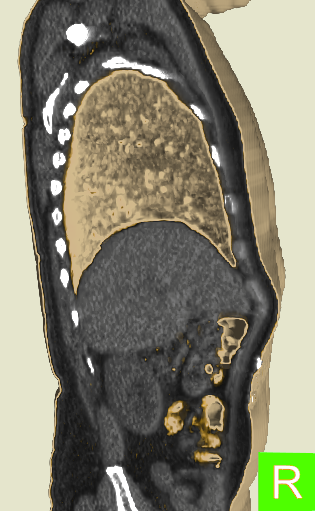}}\qquad
    \subfloat[659 ml\label{sfig:}]{\includegraphics[bb=0bp 0bp 315bp 511bp,clip,width=3.7cm]{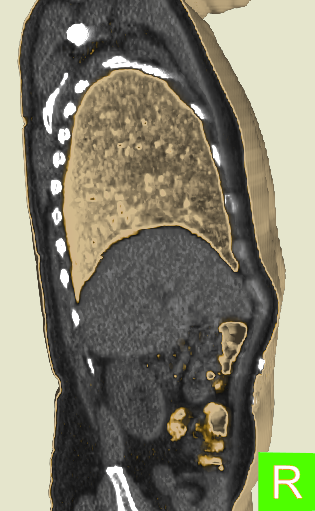}}\qquad
    \subfloat[1097 ml\label{sfig:}]{\includegraphics[bb=0bp 0bp 315bp 511bp,clip,width=3.7cm]{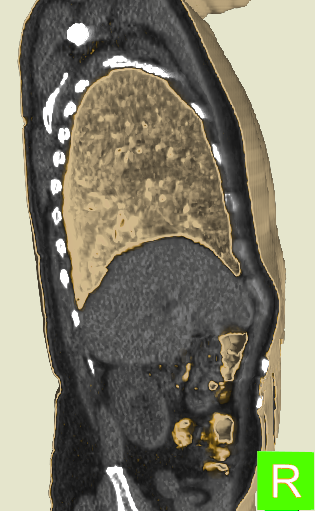}}
    \\\subfloat[0 ml\label{sfig:}]{\includegraphics[bb=0bp 0bp 315bp 511bp,clip,width=3.7cm]{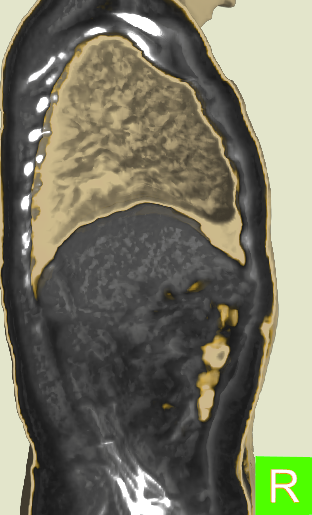}}\qquad
    \subfloat[349 ml\label{sfig:}]{\includegraphics[bb=0bp 0bp 315bp 511bp,clip,width=3.7cm]{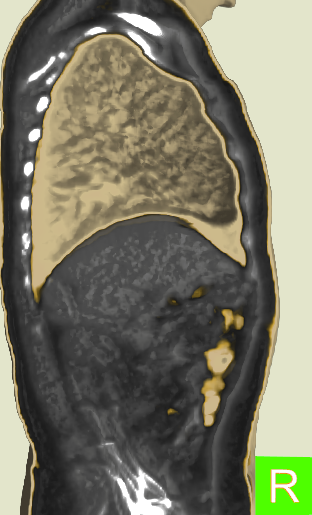}}\qquad
    \subfloat[659 ml\label{sfig:}]{\includegraphics[bb=0bp 0bp 315bp 511bp,clip,width=3.7cm]{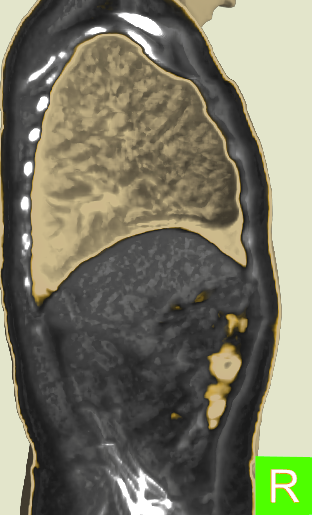}}\qquad
    \subfloat[1097 ml\label{sfig:}]{\includegraphics[bb=0bp 0bp 315bp 511bp,clip,width=3.7cm]{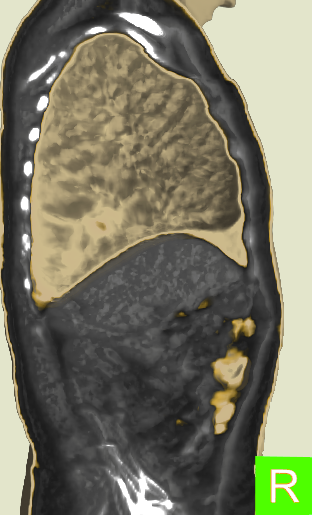}}
    \\\subfloat[0 ml\label{sfig:art}]{\includegraphics[bb=0bp 0bp 315bp 511bp,clip,width=3.7cm]{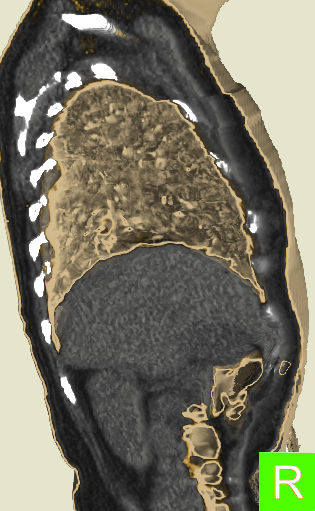}}\qquad
    \subfloat[349 ml\label{sfig:}]{\includegraphics[bb=0bp 0bp 315bp 511bp,clip,width=3.7cm]{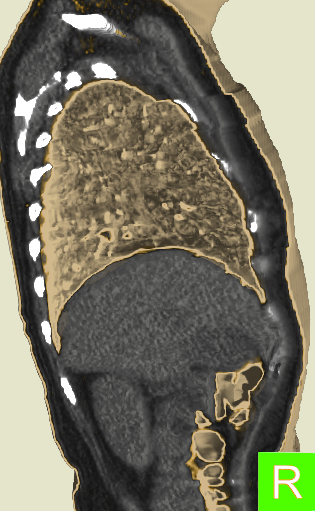}}\qquad
    \subfloat[659 ml\label{sfig:}]{\includegraphics[bb=0bp 0bp 315bp 511bp,clip,width=3.7cm]{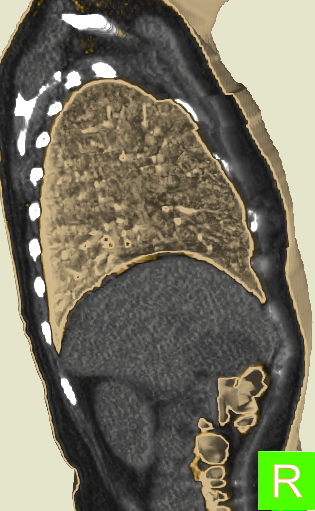}}\qquad
    \subfloat[1097 ml\label{sfig:}]{\includegraphics[bb=0bp 0bp 315bp 511bp,clip,width=3.7cm]{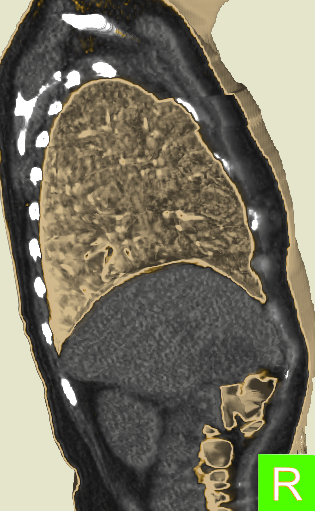}}
	\caption{3D renderings of four relative breathing volume [ml] states with coronal clipping of the (a-d) animated invididual patient (use case 1), (e-h) synthetic 4D intensity and motion atlas (use case 2) and (i-l) animated 3D patient by the 4D  motion atlas and a respiratory signal from the new 3D patient (use case 3). Cavities and skin surfaces are rendered in beige. 
}
\label{SPIE18-img:motionct}
\end{figure*}

\begin{figure}
\subfloat[Liver]{\includegraphics[bb=0bp 0bp 377bp 314bp,clip,width=0.33\textwidth]{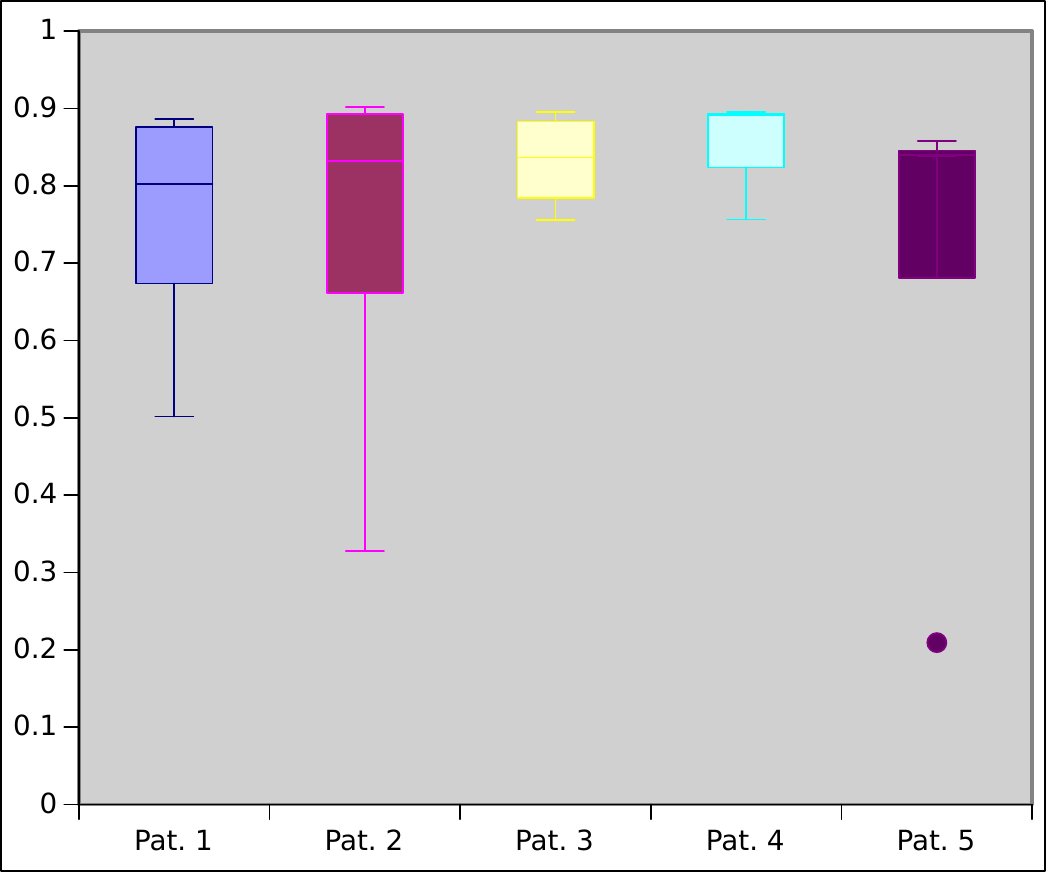}

}\subfloat[Right lung]{\includegraphics[bb=0bp 0bp 377bp 314bp,clip,width=0.33\textwidth]{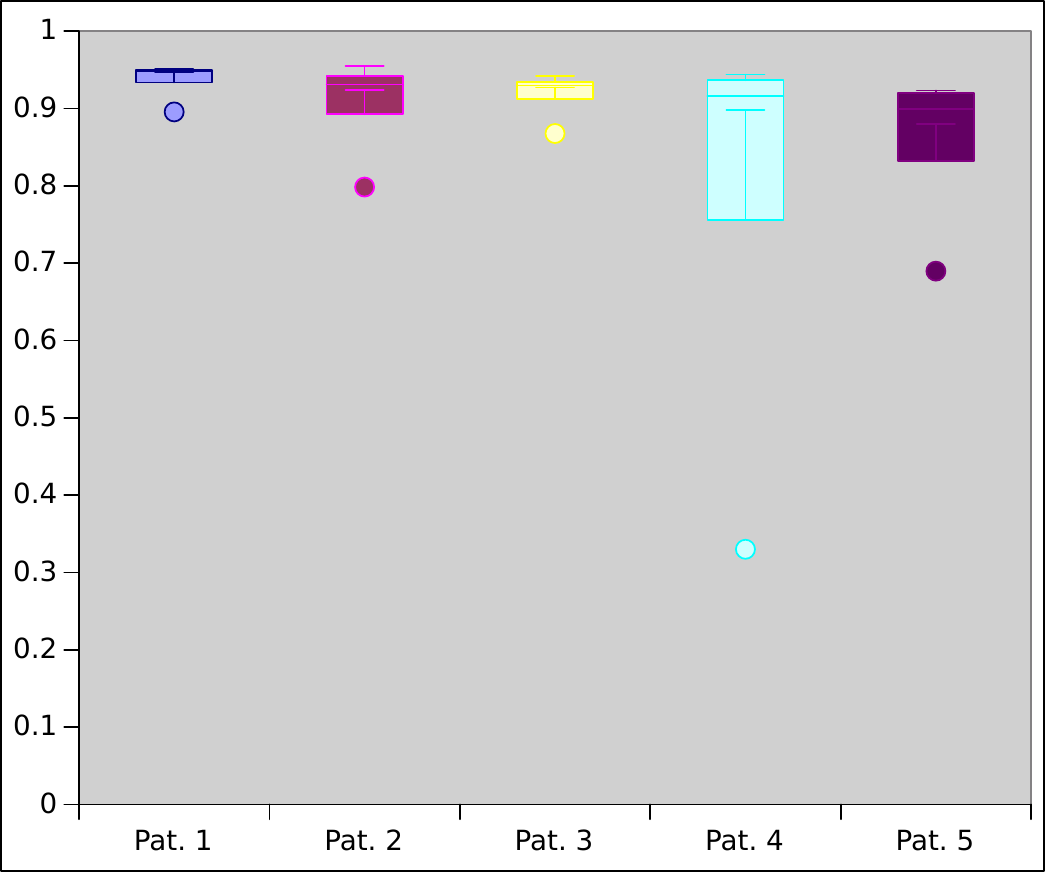}

}\subfloat[Left lung]{\includegraphics[bb=0bp 0bp 377bp 314bp,clip,width=0.33\textwidth]{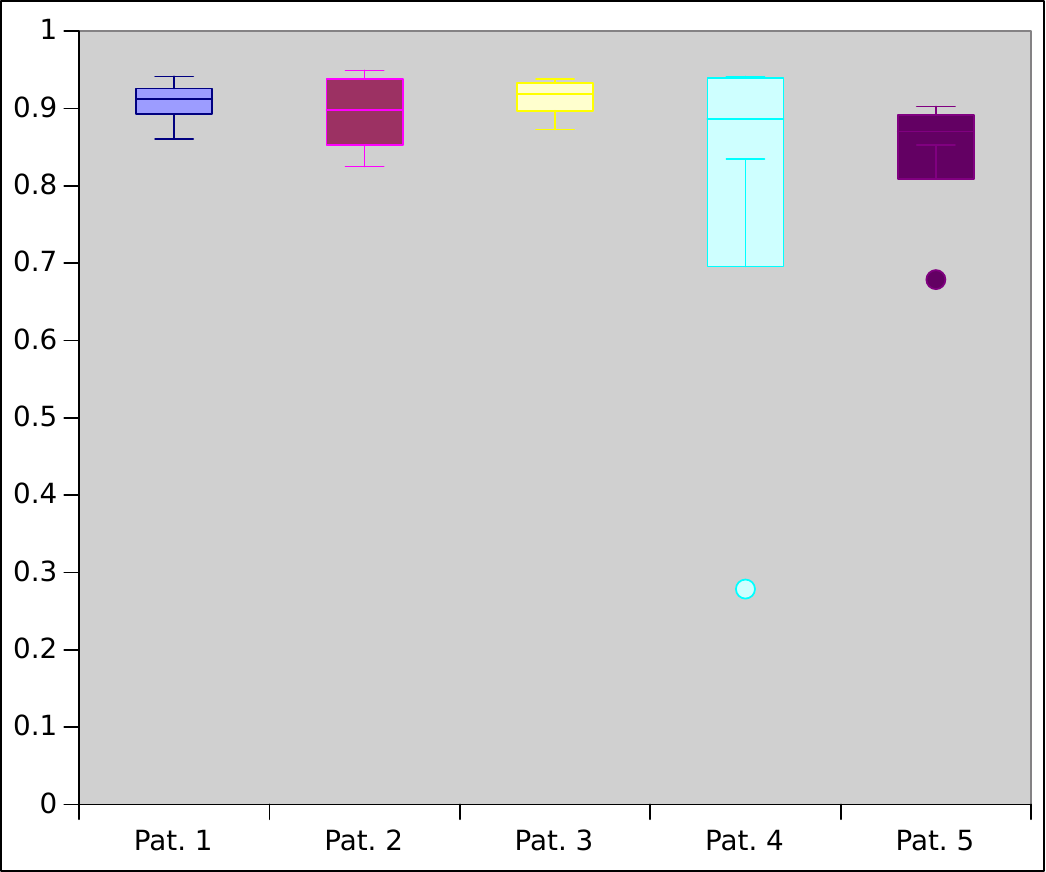}

}
\caption{\label{SPIE18-img:DSC}Inter-patient registration DICE results from a leave-one-out study for expert segmentation masks: Inter-patient (vs. intra-patient) registration still poses problems as can be seen by sporadic outliers.}
\end{figure}

\section{Results and Conclusion}

In the puncture-relevant liver region, the patients' breathing states are simulated plausibly for an individual patient (Fig. \ref {SPIE18-img:motionct}a-d), a 4D intensity and motion atlas (Fig. \ref {SPIE18-img:motionct}e-h)\footnote{\href{https://goo.gl/Qog138}{Demo movie of 4D intensity and motion atlas: https://goo.gl/Qog138}} and an animated static new 3D patient (Figs. \ref {SPIE18-img:motionct}i-l).
Cross-validation DICE values of atlas registrations of liver and lung masks to a left-out static patient yield overall median values of 0.87 (liver), 0.93 (right lung) and 0.93 (left lung). Sporadic bad inter-patient registration outliers are possible in the current process as can be seen in Fig. \ref{SPIE18-img:DSC}a-c by the dots. Here by the averaged motion, they cause artifacts at bony structures of the hip$^*$. Their influence to the process should be mitigated by a selection strategy in future work.
Statistical 4D breathing motion models for the lungs have been introduced in \cite{SPIE18-16}. Here, we use a population of 4D patient image data sets and learn a 4D intensity and motion atlas parametrized by a surrogate spirometry signal. The surrogate signal for the calculation of a mean motion model in this work was simply found by averaging. Among other obvious applications, such motion model can be warped to new 3D patient data. The animated atlas or the 3D patient can be used in a 4D VR training system with promising outcomes \cite{SPIE18-01a}. Despite the efficient regression calculus, future work will cover the direct joint transfer of the motion model coefficients.

\bibliographystyle{spiebib}

\end{document}